\documentclass[conference]{IEEEtran}
\IEEEoverridecommandlockouts
\usepackage{cite}
\usepackage{amsmath,amssymb,amsfonts}
\usepackage{graphicx}
\usepackage{textcomp}
\usepackage{xcolor}
\usepackage{cite}
\usepackage{amsmath,amssymb,amsfonts}
\usepackage{graphicx}
\usepackage{xcolor}
\usepackage{booktabs}
\usepackage{multirow}
\usepackage{stfloats}
\usepackage{array}
\usepackage{url}
\usepackage[hidelinks]{hyperref}
\usepackage{balance}
\usepackage{amsmath}  
\usepackage{amssymb}  
\usepackage{amsfonts} 
\usepackage{booktabs} 
\usepackage{tabularx}
\usepackage{tcolorbox} 
\usepackage{float}
\usepackage{algorithm}
\usepackage{algpseudocode}
\usepackage{enumitem}
\setlist{noitemsep, topsep=2pt}

\def\BibTeX{{\rm B\kern-.05em{\sc i\kern-.025em b}\kern-.08em
    T\kern-.1667em\lower.7ex\hbox{E}\kern-.125emX}}
    
\begin{document}

\title{When Confidence Fails: 
Overconfidence in LLMs under Uncertainty and Missing Clinical Information
}

\newcommand*{\affaddr}[1]{#1}
\newcommand*{\affmark}[1][*]
{\textsuperscript{#1}}
\author{
\IEEEauthorblockN{
Maryam Tahermazandarani,
Adnan Mahmood,
Fahmida Islam,
Quan Z. Sheng
}

\IEEEauthorblockA{
School of Computing, Macquarie University, Sydney, NSW 2109, Australia
}
}

\maketitle

\begin{abstract}

Large Language Models (LLMs) have recently achieved strong performance in medical question answering and clinical reasoning tasks. However, their reliability under uncertainty remains poorly understood which raises critical concerns for deployment in high-stakes clinical settings. In such environments, incorrect predictions are inherently risky, but confident incorrect predictions can be particularly harmful as they may mislead clinical decision-making. In this paper, we conduct a systematic behavioral analysis of LLMs under clinical information uncertainty. We propose an evaluation framework based on the MedMCQA dataset consisting of two complementary uncertainty settings. First, we introduce linguistic uncertainty cues through prompt modifications to simulate ambiguous clinical contexts. Second, we construct an answer removal setting, wherein the correct option is deliberately excluded mandating the model to recognize insufficient information and abstain.
We analyze both model accuracy and confidence behavior using multiple calibration metrics including calibration gap, Expected Calibration Error (ECE), and Unsafe Confident Error Rate (UCER) across 500 medical questions. Our results reveal a consistent failure mode, i.e., although accuracy degrades under increasing uncertainty, model confidence remains misaligned with accuracy. This leads to a substantial increase in unsafe confident errors, indicating that model confidence remains largely insensitive to clinically meaningful information loss.
Furthermore, we observe significant variation across models in their ability to abstain when the correct answer is unavailable, with some models persistently producing high confidence hallucinated answers. These findings expose critical limitations in the epistemic reliability of current LLMs and highlight the need for uncertainty aware evaluation methods prior to their deployment in clinical workflows.

\end{abstract}


\section{Introduction}
\label{sec:intro}
Large Language Models (LLMs) have rapidly advanced their capabilities across natural language processing tasks, i.e., reasoning and knowledge retrieval~\cite{brown2020language,10.1007/s10115-025-02620-1}. Their rapid progress has led to their adaptation in the medical domain, wherein they have achieved competitive results on challenging benchmarks, e.g., USMLE~\cite{kung2023performance} and MedMCQA~\cite{pal2022medmcqa}. This has thus resulted in an increased interest in utilizing LLMs as decision-support tools within the medical domain~\cite{singhal2023large}.
Despite these promising advancements, concerns remain regarding their reliability. LLMs are known to produce answers that appear convincing even when the underlying information is incorrect. This phenomenon, often referred to as overconfidence, occurs when LLMs provide linguistically plausible but factually incorrect answers with high certainty~\cite{ji2023survey, tripathi2025confidence}. In medicine, this is not only a technical limitation but also a safety risk that can mislead clinicians and affect patient care~\cite{steyvers2025large}. 
A key factor in a model's reliability is calibration that measures how a model's predicted confidence aligns with its accuracy~\cite{guo2017calibration}. Ideally, predictions expressed with higher confidence should correspond to higher probabilities of being correct. While earlier studies suggest that scaling models can improve their ability to accurately assess their own knowledge~\cite{kadavath2022language}, recent studies demonstrate that this is often more complex in practice.

\begin{figure}[t]
\centering
\includegraphics[width=0.65\columnwidth, trim=40 48 40 40, clip]{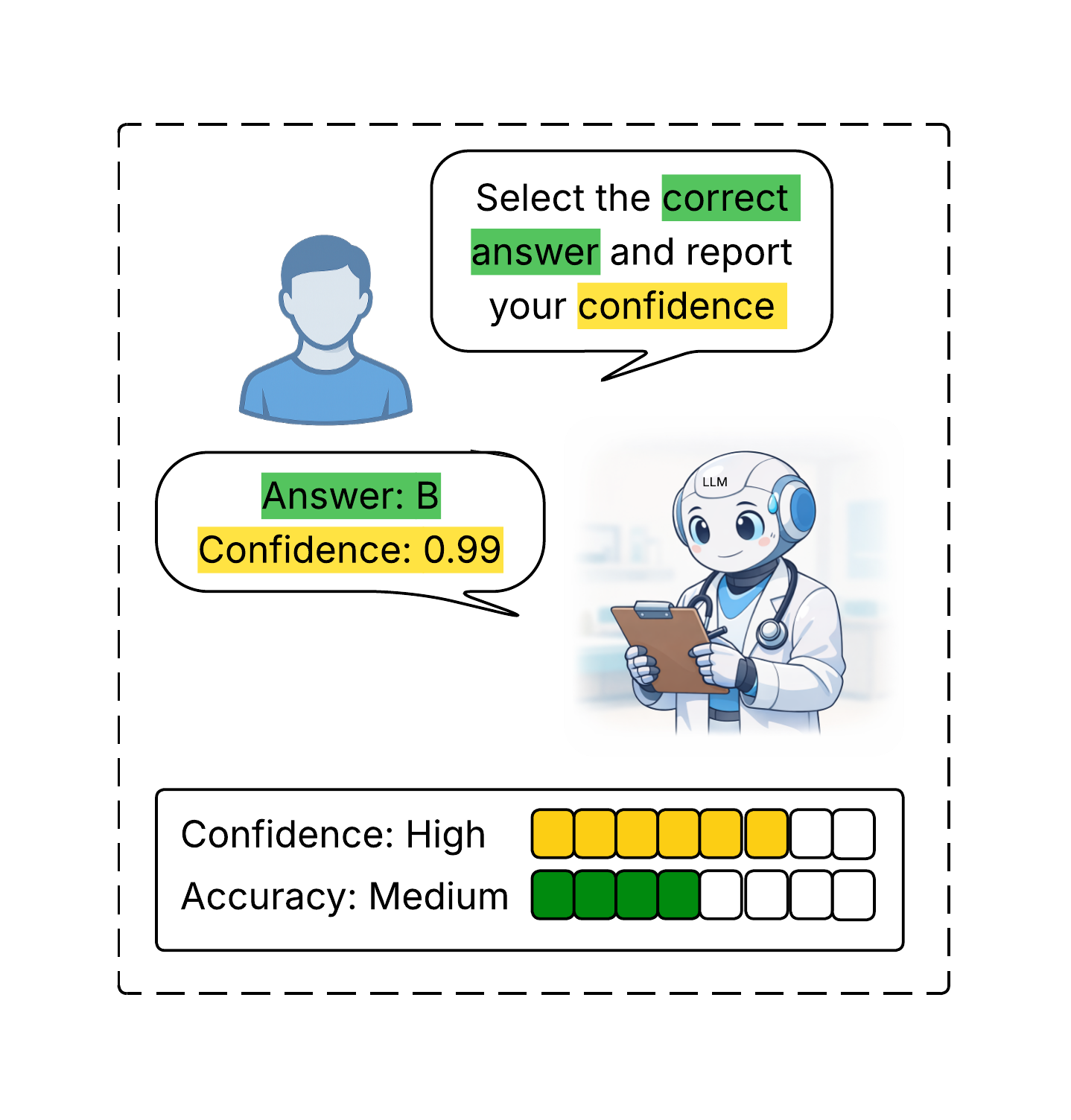}
\vspace{-4mm}
\caption{A Conceptual Illustration of Overconfidence in LLMs.}
\label{fig:mismatch}
\vspace{-5mm}
\end{figure}

LLMs can exhibit human cognitive biases, particularly when faced with ambiguous or incomplete information~\cite{xu2025language}. Moreover, the expressed confidence in generated text is not aligned with a model's internal probabilistic confidence, thereby creating a mismatch between verbalized confidence and the model's underlying beliefs~\cite{tripathi2025confidence}.
Uncertainty is a fundamental aspect of clinical reasoning. In real clinical settings, a patient's history is often incomplete and clinical signs can be ambiguous. However, most existing medical Question Answering (QA) benchmarks operate under a closed-world assumption wherein the correct answer is always among the candidate options. While this assumption simplifies evaluation, it does not reflect real clinical settings. In many scenarios, recognizing insufficient information and refraining from providing a definitive answer is the safest and most responsible course of action. While some recent studies have started to explore LLMs overconfidence in medical QA~\cite{bentegeac2025token}, there is still limited understanding of how LLMs behave when clinical information is incomplete or uncertain. This mismatch between accuracy and confidence is conceptually illustrated in Fig.~\ref{fig:mismatch}.

To address this gap, we introduce an evaluation framework to systematically investigate overconfidence in LLMs under information uncertainty. Our approach includes two complementary experimental settings, i.e., (a) uncertainty framing, wherein linguistic cues that introduce ambiguity are injected into clinical scenarios to make them uncertain and ambiguous, and (b) answer removal, wherein the correct option is removed from the candidate set and replaced with an abstention choice. These settings allow us to evaluate whether models can recognize when the available evidence is insufficient to support a confident answer. We evaluate several state-of-the-art LLMs by employing this framework and analyze their behavior using metrics, i.e., Expected Calibration Error (ECE) and Unsafe Confident Error Rate (UCER). While predictive accuracy is commonly used to evaluate medical QA systems, it does not capture whether models appropriately adjust their confidence levels when clinical information becomes incomplete or ambiguous. 
Accordingly, the salient contributions of this work are as follows:
\begin{enumerate}
 
  \item We design a dual experimental setting that captures two complementary sources of uncertainty: (i) uncertainty framing using graded linguistic cues (C0–C3), and (ii) missing information induced by removing the correct answer option and replacing it with an explicit abstention option. This setup enables systematic evaluation of model behavior under both ambiguous and insufficient evidence.
  \item We conduct a comprehensive empirical study across both proprietary and open-weight LLMs, using accuracy, calibration gap, ECE, and UCER.
  \item We identify a systematic failure of uncertainty awareness since LLMs remain highly confident under uncertainty and fail to abstain when evidence is insufficient, thus revealing a decoupling between accuracy and confidence.

\end{enumerate}

The remainder of this paper is organized as follows. Section~\ref{sec:related} reviews the related works. Section~\ref{sec:method} introduces the proposed evaluation framework followed by the experimental setup in Section~\ref{sec:setup}. Section~\ref{sec:results} presents the empirical results. Section~\ref{sec:discussion} analyzes the findings and their implications for reliable medical AI. Section~\ref{sec:limitations} discusses the limitations of the current study. Finally, Section~\ref{sec:conclusion} concludes the paper and outlines directions for future work.

\section{State of the Art}
\label{sec:related}
Since the use of LLMs is increasing in real world decision-making scenarios, assessing their reliability has become an important research direction for researchers in both academic and industry. One crucial aspect of models reliability is their ability to estimate and express their uncertainty. Recent studies suggest that despite high performance, LLMs produce hallucinated or incorrect responses with high confidence, posing significant challenges for reliable confidence estimation and calibration~\cite{huang2025survey, ji2023survey}. This issue is particularly critical in high-stakes domains, e.g., healthcare, wherein incorrect yet confident predictions may lead to misleading outcomes. Prior studies also argue that LLMs often exhibit systematic overconfidence in turn producing highly confident predictions even when they are incorrect. Behavioral analyses demonstrate that this phenomenon resembles human cognitive biases in confidence estimation~\cite{tripathi2025confidence}. To address this issue, recent studies have further explored methods for eliciting calibrated confidence from LLMs, i.e., prompting strategies and post-hoc calibration techniques~\cite{tian2023just}. However, empirical evidence indicates that verbalized confidence reported by LLMs is often poorly calibrated and may not reliably reflect a model’s internal uncertainty~\cite{groot2024overconfidence}.

A growing body of research has focused on uncertainty estimation in LLMs. In early methods, token-level probabilities derived from a model's output distribution were used to determine confidence levels. More recent works propose response level methods. For instance, Qiu et al.~\cite{qiu2024semantic} measure uncertainty by analyzing the consistency of multiple generated responses in semantic embedding space. Similarly, Wightman et al.~\cite{wightman2023strength} use disagreement among multiple outputs as an indicator of uncertainty instead of calculating just one output uncertainty. Another line of research investigates whether LLMs can recognize when they lack sufficient knowledge and should abstain from answering. For instance, Madhusudhan et al.~\cite{madhusudhan2025llms} introduce a benchmark for evaluating abstention behavior, demonstrating that even advanced models frequently fail to abstain when appropriate. These approaches focus on decision level uncertainty, i.e., whether a model chooses to produce an answer or withhold it. While prior work suggests that LLMs may exhibit limited self-awareness regarding their uncertainty~\cite{yin2023large}, their behavior under conditions of incomplete or insufficient input information remains relatively underexplored. In particular, little is known about how LLMs respond when uncertainty arises not only from model limitations, but from missing or ambiguous input information. This motivates the need for systematic evaluation of models' reliability under conditions of uncertainty and insufficient clinical evidence.

\subsection{Confidence and Overconfidence in LLMs}
Confidence in machine learning models refers to the estimated likelihood that a model’s prediction is correct. A model is considered well-calibrated when its confidence scores align with empirical accuracy, such that predictions made with probability $p$ are correct approximately $p$ of the time~\cite{guo2017calibration}. Despite significant improvements in predictive performance, modern neural networks are often poorly calibrated and tend to exhibit overconfidence, assigning higher confidence to predictions than their true correctness would justify. In the context of LLMs, confidence estimation becomes even more challenging due to their generative nature. Unlike traditional classification models with a fixed output space, LLMs operate over a vast and semantically diverse set of possible outputs, where multiple valid responses may correspond to the same input~\cite{geng2024survey}. Furthermore, LLMs express confidence in multiple ways including implicit linguistic cues as well as explicit self-reported scores, introducing additional variability in how uncertainty is represented.

Overconfidence refers to the systematic tendency of models to assign high confidence to incorrect predictions, i.e., when predicted confidence exceeds empirical accuracy~\cite{guo2017calibration, chhikara2025mind}. In high-stakes domains, e.g., in healthcare, overconfidence may appear as a form of confident failure, where a model provides incorrect medical advice with high certainty. Such responses can appear trustworthy, potentially misleading users and leading to unsafe clinical decisions~\cite{machcha2026knowing}.

\section{Methodology}
\label{sec:method}
We investigate whether LLMs exhibit overconfidence when answering clinical multiple-choice questions under uncertain or incomplete clinical information. We introduce two experimental settings that simulate uncertainty in clinical decision-making, thereby enabling a systematic analysis of a model's behavior under uncertainty.

\begin{figure}[t]
\centering
\includegraphics[width=\columnwidth, trim=0 56 0 56, clip]{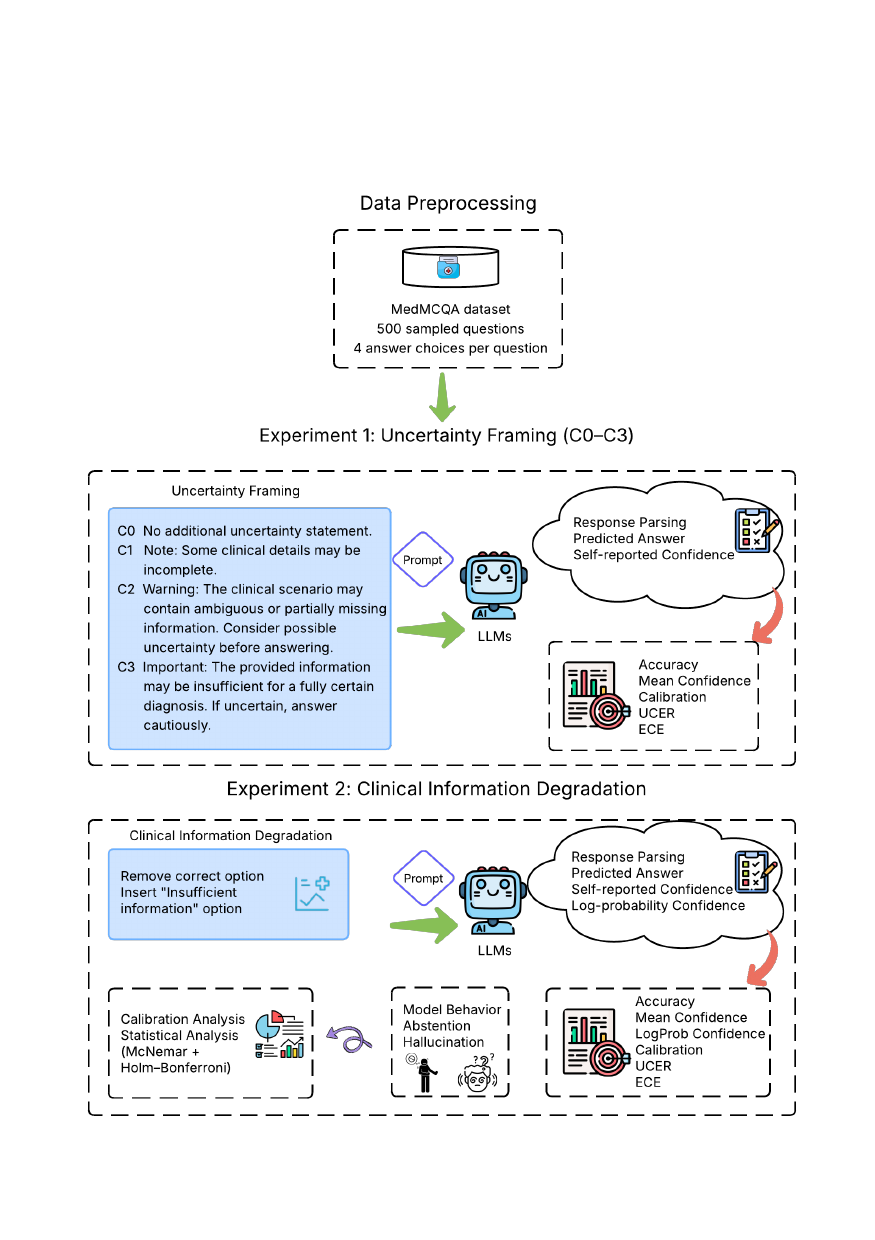}
\vspace{-4mm}
\caption{An Overview of the Proposed Evaluation Framework for Assessing
Overconfidence in LLMs.}
\label{fig:framework}
\vspace{-5mm}
\end{figure}

\subsection{Problem Formulation}
Let $\mathcal{D} = \{(x_i, A_i, y_i)\}_{i=1}^n$ denote a medical multiple-choice dataset consisting of $n$ samples. Each sample includes a clinical question $x_i$, a set of candidate answer options $A_i = \{a_{i,1} , a_{i,2}, \dots, a_{i,K}\}$ with $K$ being the number of candidate options, and a ground-truth label $y_i$ in $A_i$.
Given an instance $(x_i, A_i)$, a LLM parameterized by $\theta$, i.e., $f_\theta$, outputs a predicted answer $\hat{y}_i \in A_i$ along with an associated confidence score $c_i \in [0, 1]$ as:
\begin{equation}
(\hat{y}_i, c_i) = f_\theta(x_i, A_i)
\end{equation}
%
where, $\hat{y}_i$ represents the selected answer option and $c_i$ denotes the confidence reported by the model.

Ideally, a model is well-calibrated when its predicted confidence reflects the true probability of correctness. In other words, the empirical accuracy of predictions with confidence $p$ should equal $p$. Formally, for any $p \in [0,1]$, a calibrated model satisfies:
\begin{equation}
P(\hat{y}_i = y_i \mid c_i = p) = p
\end{equation}
%
where, $y_i$ is the ground-truth answer. Deviations from this condition indicate miscalibration, particularly, when a model produces incorrect prediction with high confidence.
Our objective is to evaluate whether this calibration property holds when clinical information becomes uncertain or when a correct answer is unavailable.

\subsection{Uncertainty Framing}
The first setting ascertains whether the linguistic expressions of uncertainty influence model confidence and calibration. In clinical practice, decision-making often occurs under incomplete or ambiguous clinical information. It is, therefore, highly indispensable to assess whether LLMs appropriately adjust their confidence when uncertainty is explicitly introduced in the prompt. We introduce linguistic cues that signal potential ambiguity or incompleteness. 
Let $x_i^{(k)}$ denote a prompt variant corresponding to uncertainty level $k$, defined as:
\begin{equation}
x_i^{(k)} = g_k(x_i)
\end{equation}
%
where, $g_k(\cdot)$ represents a transformation that injects contextual uncertainty while preserving the original clinical content. We develop four variants as depicted in Table~\ref{tab:table1}.
Importantly, the answer options $A_i$ and ground-truth label $y_i$ remain unchanged across all framing conditions. This design ensures that the underlying clinical reasoning task remains identical, thereby allowing us to isolate the effect of uncertainty framing on a model's confidence and calibration.

\subsection{Answer Removal}
The second experimental setting evaluates model behavior when the correct answer is unavailable. This setting reflects clinical scenarios in which available evidence is insufficient to support a definitive diagnosis or treatment decision.
In this setting, the correct answer $y_i$ is removed from the candidate set $A_i$ and replaced with an abstention option:
\begin{equation}
A'_i = (A_i \setminus \{y_i\}) \cup \{a_{\text{insufficient}}\}
\end{equation}

\noindent where, $a_{\text{insufficient}}$ corresponds to the option ``Insufficient information to determine the correct answer''. Selecting $a_{\text{insufficient}}$ represents a safe abstention. We define a hallucinated prediction as any output $\hat{y}_i \in A'_i \setminus \{a_{\text{insufficient}}\}$, since the correct answer is no longer present in the candidate set.
 
This setting enables a direct evaluation of whether models can recognize missing evidence and abstain appropriately, or instead generate unsupported answers with high confidence. For each prediction, we collect two complementary confidence measures:

\begin{itemize}
    \item Self-reported ($c_i$): The explicit confidence score provided by a model in its output; and
\item Log probability ($p_{\hat{y}_i}$): The probability of each answer option is computed by aggregating token-level logits corresponding to the option
identifier, e.g., A, B, C, and D
in a model's output.
\end{itemize}

Self-reported confidence is obtained by prompting a model to provide an explicit numerical confidence value alongside its predicted answer. Log probability confidence is computed from the token-level logits and normalized across candidate answer options. 
This measure reflects a model’s internal probabilistic belief and enables direct comparison between explicitly reported confidence and implicit probability estimates derived from a model’s output distribution.

\begin{table}[t]
\centering
\caption{Uncertainty Framing Statements.}
\label{tab:table1}
\begin{tabular}{p{2cm} p{6cm}}
\toprule
\textbf{Condition} & \textbf{Framing Statement} \\
\midrule
C0 & No additional uncertainty statement. \\

C1 & Note: Some clinical details may be incomplete. \\

C2 & Warning: The clinical scenario may contain ambiguous or partially missing information. Consider possible uncertainty before answering. \\

C3 & Important: The provided information may be insufficient for a fully certain diagnosis. If uncertain, answer cautiously. \\
\bottomrule
\end{tabular}
\end{table}

\subsection{Evaluation Metrics}
\label{sec:metrics}
We evaluate a model's behavior along three key dimensions, i.e., predictive accuracy, confidence calibration, and safety under uncertainty. 

\subsubsection{Calibration and Probability Metrics}

To quantify the alignment between predicted confidence and empirical correctness, we compute ECE following~\cite{guo2017calibration}. The confidence interval $[0,1]$ is partitioned into $M=10$ equal-width bins. ECE is defined as:
\begin{equation}
\mathrm{ECE} = \sum_{m=1}^{M} \frac{|B_m|}{n}\,
\left| \mathrm{acc}(B_m) - \mathrm{conf}(B_m) \right|
\label{eq:ece}
\end{equation}
%
where, $n$ is the total number of samples, $B_m$ denotes the set of samples whose confidence falls into bin $m$, $\mathrm{acc}(B_m)$ is the average accuracy within the bin, and $\mathrm{conf}(B_m)$ is the average~confidence.

In addition to self-reported confidence ($c_i$), we compute an implicit estimate derived from token-level logits. Let $\ell_k$ denote the logit associated with the token representing answer option $k$, e.g., ``A'', ``B'', ``C'', and ``D'', at the prediction step. The logits are extracted at the decoding step where the model generates the answer token. The normalized probability $p_k$ is computed via the softmax function:
\begin{equation}
p_k = \frac{\exp(\ell_k)}{\sum_{j=1}^{K} \exp(\ell_j)}
\end{equation}
%
where, $K$ is the number of candidate options. The log probability based confidence is defined as the probability assigned to the selected answer:
\begin{equation}
\mathrm{Conf}_{\text{logprob}} = p_{\hat{y}_i}
\end{equation}
%
where, $\hat{y}_i$ denotes the predicted answer option for sample $i$.

This metric enables comparison between a model's implicit probabilistic confidence and its explicitly reported confidence.

\subsubsection{Safety and Behavioral Metrics}
To quantify high-confidence errors, we employ the UCER metric defined as:
\begin{equation}
\mathrm{UCER}_{\tau} = \frac{1}{n}\sum_{i=1}^{n}
\mathbb{I}\!\left[\hat{y}_i \neq y_i \ \land\ c_i \ge \tau\right]
\label{eq:ucer}
\end{equation}
%
where, $\hat{y}_i$ is the predicted answer, $y_i$ is the ground-truth label, $\tau$ is a high-confidence threshold, and $\mathbb{I}[\cdot]$ denotes the indicator function.
This metric captures incorrect predictions made with high confidence and is particularly critical in safety-sensitive settings like clinical decision-making. UCER directly captures unsafe model behavior, where incorrect predictions are accompanied by unjustified high confidence.

In the answer removal setting, selecting the option $a_{\text{insufficient}}$ is interpreted as an abstention. Conversely, any prediction $\hat{y}_i \in A'_i \setminus \{a_{\text{insufficient}}\}$ is defined as a hallucinated prediction, since the correct answer is no longer present in the candidate set. We compute the corresponding rates as the fraction of samples satisfying each condition:
\begin{equation}
\mathrm{Abstention} =
\frac{1}{n}
\sum_{i=1}^{n}
\mathbb{I}[\hat{y}_i = a_{\text{insufficient}}]
\end{equation}
\begin{equation}
\mathrm{Hallucination} =
1 - \mathrm{Abstention}
\end{equation}

\subsubsection{Statistical Analysis}

Given that all models are evaluated on the same set of questions, we employ paired statistical tests. For binary outcomes, we apply McNemar’s~\cite{mcnemar1947note} test to assess differences in model accuracy. For comparisons of self-reported confidence, we apply the Wilcoxon~\cite{wilcoxon1945individual} signed-rank test as confidence values are bounded in $[0,1]$ and may not follow a normal distribution. All hypothesis tests were two-sided with a significance threshold of $\alpha = 0.05$ (5\% level), a commonly adopted threshold for controlling Type I error in statistical hypothesis testing. To control for multiple comparisons in pairwise statistical testing, we applied the Holm–Bonferroni correction~\cite{holm1979simple}.

\section{Experimental Setup}
\label{sec:setup}

\subsection{Dataset and Sampling}
We utilized MedMCQA \cite{pal2022medmcqa} a large-scale multiple-choice benchmark derived from Indian medical entrance examinations. The dataset comprises questions across diverse medical subjects and reasoning types. From the full dataset, we randomly sampled $500$ single answer multiple-choice questions using a fixed random seed to ensure reproducibility. Only questions with four answer options and a clearly defined correct label were retained. Each question was treated independently.

\subsection{Implementation of Experimental Settings}

Following the framework described in Section~\ref{sec:method}, we evaluated model behavior under two experimental settings.
In the first setting, we constructed four prompt variants ($C_0$--$C_3$). This setup resulted in a total of $2,000$ evaluation instances, i.e., $500$ questions evaluated under four framing conditions.
In the second experiment, the ground-truth option was removed and replaced with an abstention option, ``Insufficient information to determine the correct answer''. It is pertinent to note that the position of this option is randomized for each question to mitigate positional bias. 

\subsection{Models and Inference Protocol}
\begin{figure}[H]
\centering
\includegraphics[width=0.9\columnwidth, trim=15 86 25 95, clip ]{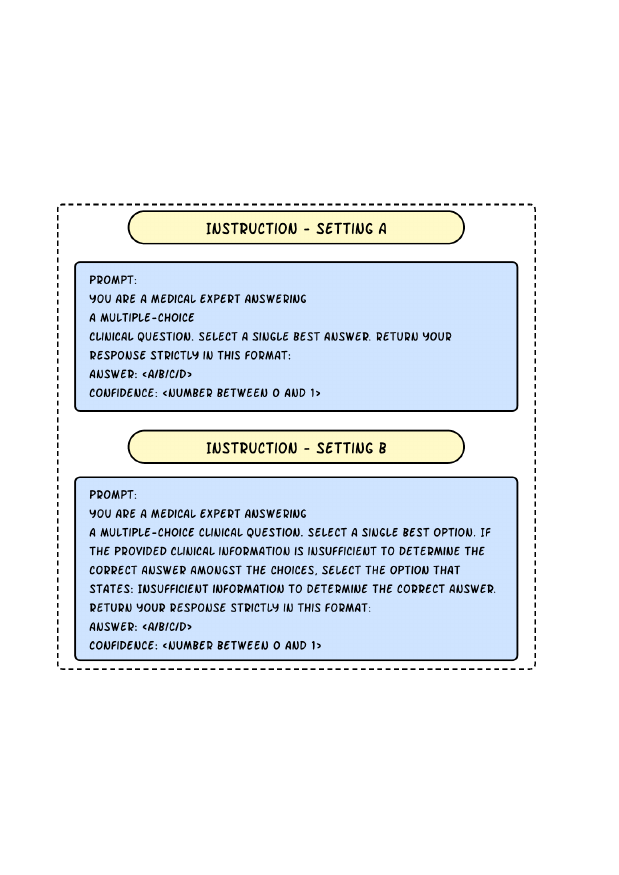}
\vspace{-4mm}
\caption{Two Prompt Settings Used in Our Experiments: (A) Standard Answer Selection 
and (B) Answer Selection with an Explicit Insufficient Information Option.}
\label{fig:prompt}
\vspace{-3mm}
\end{figure}

We evaluated five LLMs comprising both proprietary and open-weight models, i.e., GPT-5, GPT-4o, GPT-4o-mini, LLaMA-3-8B-Instruct, and Mistral-7B-Instruct-v0.3. The selection enabled comparisons across model families with different training paradigms and parameter scales. GPT-5, GPT-4o, and GPT-4o-mini were accessed through API endpoints, whereas, LLaMA-3-8B-Instruct and Mistral-7B-Instruct-v0.3 were run locally via the Hugging Face Transformers library with 4-bit quantization on an NVIDIA T4 GPU. All inference was conducted with a temperature set to $0$ to ensure deterministic and reproducible outputs. We employed a structured prompting strategy with a fixed instruction format across all experiments, thereby requiring models to produce both a discrete answer and an associated confidence score. This design enables consistent evaluation of model calibration and uncertainty behavior. The full prompt templates are depicted in Fig.~\ref{fig:prompt}.
Model performance was evaluated using the metrics defined in Section~\ref{sec:metrics}.

\section{Results}
\label{sec:results}

\subsection{Performance under Uncertainty Framing (Setting A)}

\begin{figure*}[!t]
\includegraphics[width=\textwidth , trim=0 5 0 5, clip]{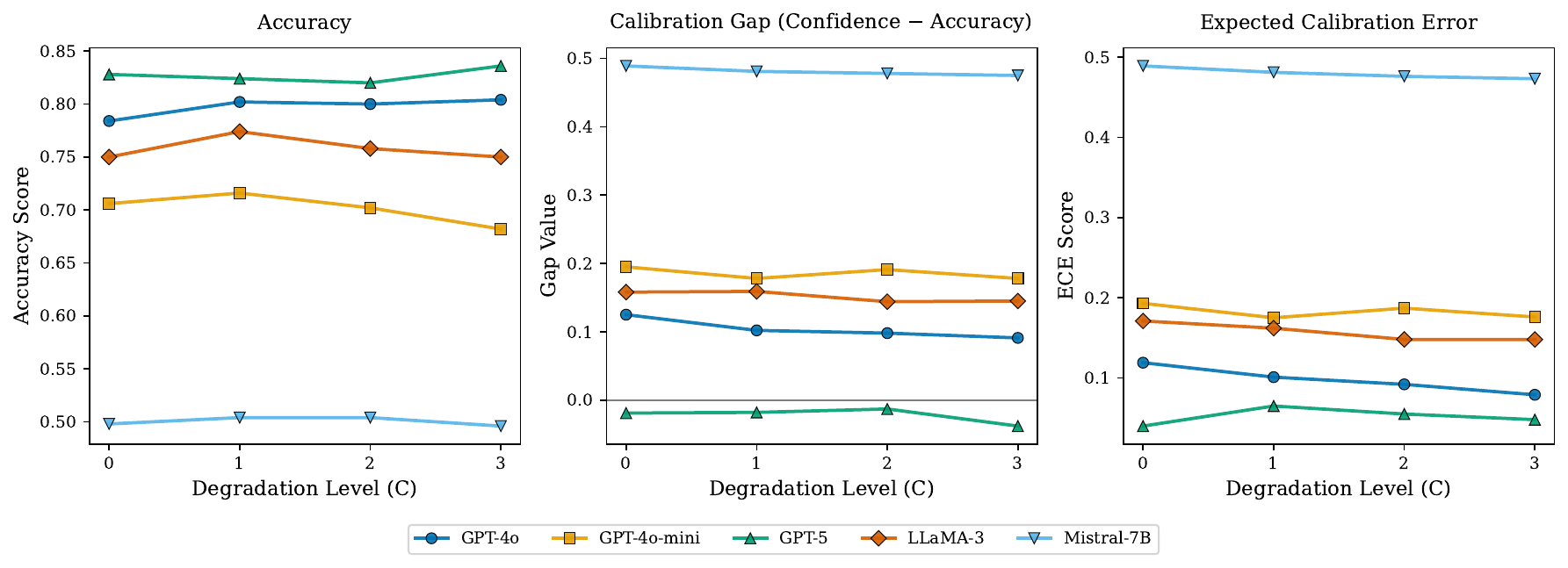}
\caption{Accuracy and Self-Overconfidence Gap (Defined as the Difference Between Mean Self-Reported Confidence and Accuracy) across Uncertainty Framing Conditions (C0–C3). Higher Conditions Correspond to Increasing Levels of Prompt Induced Uncertainty.}
\label{fig:calibration_results}
\end{figure*}

Table~\ref{tab:settingA} summarizes the models’ performance across four uncertainty framing conditions (C0–C3) whereas, Fig.~\ref{fig:calibration_results} illustrates the relationship between accuracy and calibration gap. A key observation is that predictive accuracy remains relatively stable across uncertainty conditions for all models. For instance, GPT-5 consistently achieves the highest accuracy, i.e., with values ranging from $0.82$ to $0.84$, showing minimal sensitivity to uncertainty framing. Similar trends are observed for GPT-4o and GPT-4o-mini. This indicates that introducing uncertainty cues in the prompts has a negligible impact on the answer selection behavior.

In contrast, calibration behavior differs substantially across models. GPT-5 is the only model that demonstrates comparatively well-calibrated behavior with confidence values slightly lower than accuracy hence resulting in small negative calibration gaps. This consistent underconfidence suggests that GPT-5 provides relatively conservative confidence estimates. In comparison, GPT-4o and GPT-4o-mini consistently exhibit overconfidence across all conditions, where reported confidence exceeds the empirical accuracy. Notably, this miscalibration remains largely unchanged as uncertainty cues become stronger thus indicating that these two models fail to adjust their confidence in response to increasing ambiguity in the input.
This effect is significantly amplified in open-weight models: (a) LLaMA achieves moderate accuracy of $0.75$, but maintains high confidence resulting in substantial overconfidence gaps and (b) Mistral demonstrates the most severe miscalibration with accuracy $\approx 0.50$, however, its confidence remains close to $1$ with ECE and UCER being highest amongst all models. In practice, this implies that a substantial fraction of predictions are both incorrect and highly confident, which represents a critical failure mode for deployment in safety-critical domains. 

Importantly, uncertainty framing does not lead to any meaningful reduction in confidence across any model. Even under the strongest uncertainty condition (C3) confidence levels remain consistently high. These results indicate that LLMs fail to internalize uncertainty signals introduced at the prompt level. Overall, these findings reveal a fundamental mismatch between predictive performance and confidence behavior under uncertainty. While accuracy remains stable, confidence fails to appropriately reflect uncertainty consequently leading to systematic overconfidence. In medical QA, such confidently incorrect predictions pose significant risks for clinical decision-making and highlight a key limitation in the reliability of current LLMs.

\begin{table}[t]
\centering
\caption{Performance and Calibration Metrics.}
\label{tab:settingA}
\begin{tabular}{lcccccc}
\toprule
Model & C & Acc. & Conf. & Gap & UCER & ECE \\

\midrule

\multirow{4}{*}{GPT-4o}
 & 0 & 0.784 & 0.909 & 0.125 & 0.180 & 0.119 \\
 & 1 & \textbf{0.802} & 0.904 & 0.102 & 0.136 & 0.101 \\
 & 2 & 0.800 & 0.898 & 0.098 & 0.128 & 0.092 \\
 & 3 & 0.804 & 0.895 & 0.091 & 0.130 & 0.079 \\

\midrule

\multirow{4}{*}{GPT-4o-mini}
 & 0 & 0.706 & 0.901 & 0.195 & 0.248 & 0.193 \\
 & 1 & 0.716 & 0.894 & 0.178 & 0.226 & 0.175 \\
 & 2 & 0.702 & 0.893 & 0.191 & 0.218 & 0.187 \\
 & 3 & 0.682 & 0.860 & 0.178 & 0.162 & 0.176 \\

\midrule

\multirow{4}{*}{GPT-5}
 & 0 & \textbf{0.828} & 0.809 & -0.019 & 0.034 & 0.040 \\
 & 1 & 0.824 & 0.806 & -0.018 & 0.028 & 0.065 \\
 & 2 & 0.820 & 0.807 & -0.013 & 0.030 & 0.055 \\
 & 3 & \textbf{0.836} & 0.798 & -0.038 & 0.028 & 0.048 \\

\midrule

\multirow{4}{*}{LLaMA-3}
 & 0 & 0.750 & 0.908 & 0.158 & 0.222 & 0.171 \\
 & 1 & 0.774 & 0.933 & 0.159 & 0.218 & 0.162 \\
 & 2 & 0.758 & 0.902 & 0.144 & 0.226 & 0.148 \\
 & 3 & 0.750 & 0.895 & 0.145 & 0.212 & 0.148 \\

\midrule

\multirow{4}{*}{Mistral-7B}
 & 0 & 0.498 & 0.987 & 0.489 & 0.498 & 0.489 \\
 & 1 & 0.504 & 0.985 & 0.481 & 0.492 & 0.481\\
 & 2 & 0.504 & 0.982 & 0.478 & 0.488 & 0.476 \\
 & 3 & 0.496 & 0.971 & 0.475 & 0.492 & 0.473 \\

\bottomrule
\end{tabular}
\end{table}

\subsection{Degraded Setting}
In this setting, we evaluated model behavior under explicit information insufficiency by removing the correct answer from the candidate set resulting in scenarios with no valid answer. Under such conditions a reliable and uncertainty aware model is expected to abstain rather than produce unsupported predictions. The results across models are summarized in Table~\ref{tab:experiment2_calibration} and Table~\ref{tab:experiment2_behavior}, and the overall trends are visualized in Fig.~\ref{fig:phase1} and Fig.~\ref{fig:confidence}.

\subsubsection{Abstention Performance}

\begin{figure*}[!t]
\includegraphics[width=\textwidth, trim=0 11 0 5, clip]{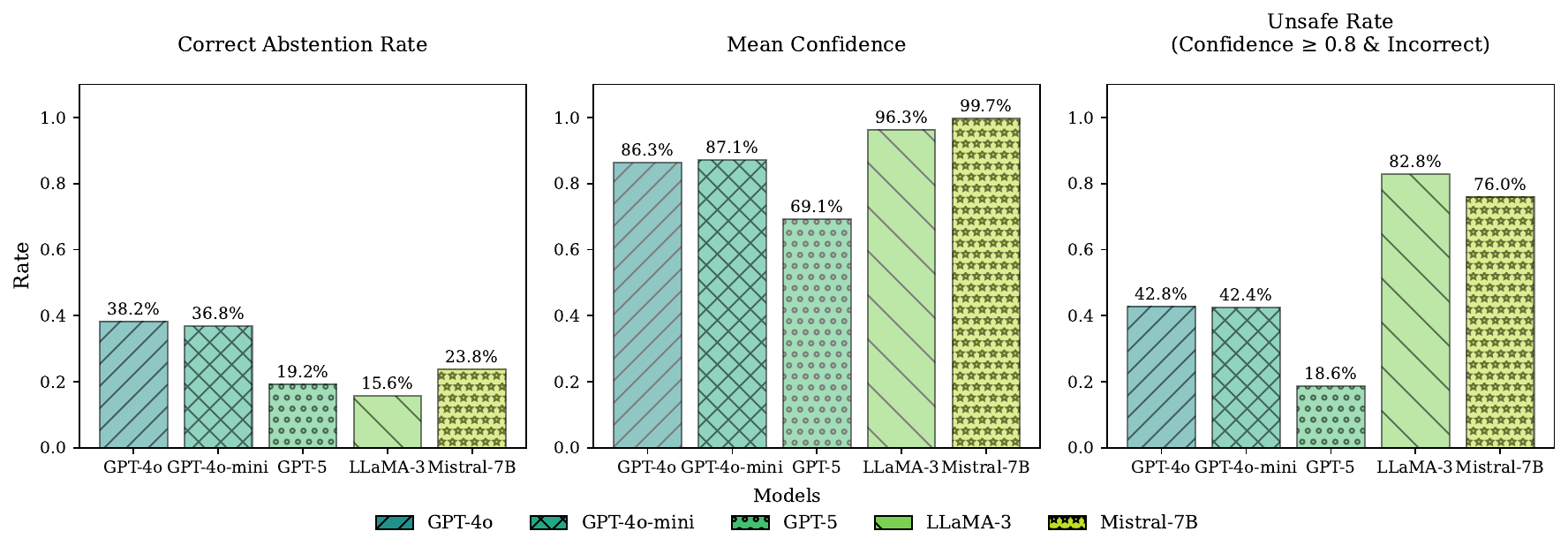}
\vspace{-4mm}
\caption{Correct Abstention Rate, Mean Confidence, and Unsafe Rate across Uncertainty Framing Conditions (C0–C3). Unsafe Predictions are Defined as Incorrect Answers with Confidence Greater Than 0.8. 
C0–C3 Represent Increasing Levels of Uncertainty in the Prompt.}
\label{fig:phase1}
\end{figure*}

Abstention performance is defined as the proportion of cases in which a model correctly selects the ``insufficient information'' option when the correct answer is unavailable. As depicted in Table~\ref{tab:experiment2_behavior} and Fig.~\ref{fig:phase1} abstention rates remain consistently low across all models. Among all models, GPT-4o achieves the highest abstention rate of $0.382$, followed closely by GPT-4o-mini with $0.368$. However, even these models abstain in fewer than $40\%$ of cases. In contrast, open-weight models demonstrate substantially lower abstention rates. LLaMA abstains in only $15.6\%$ of cases, and Mistral abstains in $23.8\%$ of cases. These findings reveal a systematic failure of uncertainty awareness in LLMs. Even when the correct answer is explicitly removed, most models continue to select one of the remaining answer options rather than abstaining. This indicates a strong bias toward forced answering, whereby models tend to select a plausible answer over explicitly acknowledging insufficient information, even when an explicit abstention option is available. One possible explanation for this behavior lies in both the next-token prediction objective of LLMs and their training distribution and instruction-following bias. Since LLMs are predominantly trained on tasks where a correct answer exists, they develop a preference for generating informative responses rather than selecting abstention options. As a result, even when an explicit ``insufficient information'' option is available, models tend to favor standard answer choices, thus resulting in systematic failures in recognizing insufficient or missing clinical information.

\begin{table}[t]
\centering
\caption{Model Calibration and Safety Metrics under Answer Removal Conditions (Experiment 2).}
\label{tab:experiment2_calibration}
\begin{tabular}{lcccccc}
\toprule
Model & Acc. & Conf. & LogProb & Gap & UCER & ECE \\
\midrule
GPT-4o & \textbf{0.382} & 0.863 & 0.883 & 0.481 & 0.428 & 0.486 \\
GPT-4o-mini & 0.368 & 0.871 & 0.913 & 0.503 & 0.424 & 0.491 \\
GPT-5 & 0.192 & 0.691 & -- & 0.499 & 0.186 & 0.600 \\
LLaMA-3 & 0.156 & 0.970 & 0.280 & \textbf{0.814} & \textbf{0.828} &\textbf{ 0.814} \\
Mistral-7B & 0.238 & \textbf{0.998} & 0.657 & 0.760 & 0.760 & 0.760 \\
\bottomrule
\end{tabular}
\end{table}

\begin{table}[t]
\centering
\caption{Model Behavior under Insufficient-information Conditions (Experiment 2).}
\label{tab:experiment2_behavior}
\begin{tabular}{lcccc}
\toprule
Model & Abstention & Hallucination & MeanConf & Unsafe \\
\midrule
GPT-4o & \textbf{0.382} & 0.618 & 0.863 & 0.428 \\
GPT-4o-mini & 0.368 & 0.632 & 0.871 & 0.424 \\
GPT-5 & 0.192 & \textbf{0.808} & 0.691 & 0.186 \\
LLaMA-3 & 0.156 & 0.798 & 0.963 & \textbf{0.828} \\
Mistral-7B & 0.238 & 0.762 & \textbf{0.997} & 0.760 \\
\bottomrule
\end{tabular}
\end{table}

\subsubsection{High-confidence Hallucinations}
\begin{figure}[!t]
\centering
\includegraphics[width=\columnwidth , trim=0 5 0 5, clip]{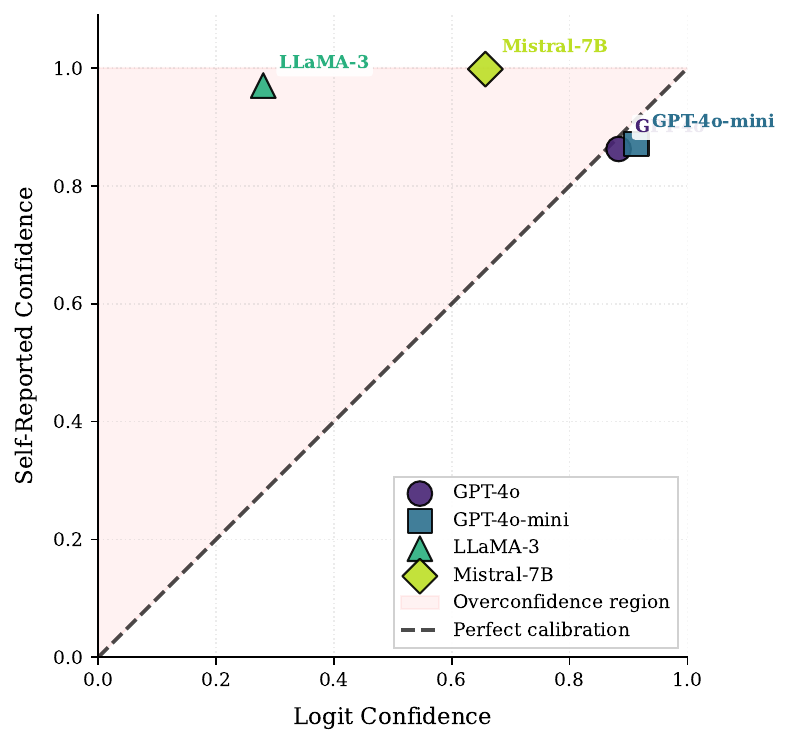}
\vspace{-4mm}
\caption{
Self-reported vs. Logit-based Confidence across the Models. The Diagonal Line Denotes Perfect Calibration, whereas, Shaded Region Indicates Overconfidence.}
\label{fig:confidence}
\vspace{-4mm}
\end{figure}

When models fail to abstain under information insufficiency, they generate hallucinated answers by selecting incorrect options. As depicted in Table~\ref{tab:experiment2_behavior} and Fig.~\ref{fig:confidence}, hallucination rates remain high across all models. GPT-5 exhibits the highest hallucination rate of $0.808$, followed by LLaMA with $0.798$ and Mistral with $0.762$. More critically, these hallucinated predictions are frequently associated with high confidence. We define unsafe predictions as incorrect answers with confidence greater than $0.8$. Under this definition LLaMA exhibits the highest unsafe rate of $0.828$, while Mistral also exhibits a high unsafe rate of $0.760$. Even stronger models, including GPT-4o and GPT-4o-mini, produce substantial unsafe predictions ($0.428$ and $0.424$, respectively) as illustrated in Fig.~\ref{fig:phase1}.

These findings reveal that hallucinated answers are frequently accompanied by high confidence, directly contributing to elevated UCER values. Notably, open-weight models exhibit the highest UCER, indicating a substantial proportion of unsafe predictions. Among all models, LLaMA exhibits the worst safety profile, characterized by the highest UCER and confidence gap. In contrast, GPT-5 exhibits the lowest UCER indicating comparatively safer behavior although hallucination rates remain high under this setting. This behavior represents a critical safety risk in clinical decision-support settings, as confidently incorrect predictions may mislead clinicians and lead to unsafe medical decisions.

\subsubsection{Self-reported vs. Logit-based Confidence}
We analyzed the relationship between self-reported confidence and logit-based probabilities derived from token-level likelihoods. As illustrated in Table~\ref{tab:experiment2_calibration}, substantial discrepancies are observed between these two measures across all models. For instance, LLaMA reports a very high average confidence of $0.970$, while its logit-based confidence remains considerably lower at $0.280$ resulting in a large confidence gap of $0.814$. Mistral exhibits a similar pattern with near maximal reported confidence of $0.998$ despite only moderate logit probabilities of $0.657$. These results indicate that self-reported confidence systematically overestimates the probability implied by a model’s internal logits. 
In contrast, GPT-based models demonstrate closer alignment between reported and logit-based confidence values suggesting relatively better calibration. However, this alignment remains imperfect and noticeable discrepancies persist. Logit-based confidence is not available for GPT-5 due to API limitations thus only self-reported confidence is considered for this model. 

Overall, these findings reveal a fundamental inconsistency between expressed and internal confidence signals. This suggests that self-reported confidence cannot be reliably interpreted as a measure of model uncertainty particularly in safety critical medical applications. Compared to setting A wherein models fail to adjust confidence under ambiguity this setting exposes a more severe limitation, models continue to produce confident predictions even when no correct answer is available. This highlights a fundamental failure in uncertainty awareness under explicit information insufficiency raising concerns about the safe deployment of LLMs in clinical settings.

\subsection{Statistical Analysis}
We also performed pairwise McNemar tests to evaluate the statistical differences in model accuracy. To control for multiple comparisons, we applied the Holm–Bonferroni correction.
As observed in Table~\ref{tab:mcnemar_results}, several model pairs exhibit statistically significant differences in accuracy with $p < 0.05$, particularly between proprietary and open-weight models.
Specifically, GPT-4o and GPT-4o-mini show significant differences compared to GPT-5, LLaMA, and Mistral. In contrast, no statistically significant difference is observed between GPT-4o and GPT-4o-mini hence indicating comparable predictive performance between these two models.
Similarly, no significant difference is found between GPT-5 and either LLaMA ($p = 0.133$) or Mistral ($p = 0.077$) suggesting that their accuracy levels are statistically indistinguishable. However, the difference between LLaMA and Mistral is statistically significant ($p = 0.002$). 

After Holm–Bonferroni correction, most significant differences remain. In particular, GPT-4o and GPT-4o-mini continue to significantly outperform the other evaluated models. In contrast, no significant differences are observed between GPT-5 and the open-weight models, indicating comparable accuracy within the open-weight group. The difference between LLaMA and Mistral also remains statistically significant after correction. Overall, these results confirm that performance differences between proprietary and open-weight models are statistically robust, while differences within each group are less pronounced. Importantly, these statistical findings align with the observed differences in calibration and safety behavior thus reinforcing the reliability of our empirical analysis.

\begin{table}[t]
\centering
\caption{Pairwise McNemar Test Results for Model Accuracy Comparison.}
\label{tab:mcnemar_results}
\begin{tabular}{lcc}
\toprule
Model Pair & $p$-value & Significant ($p<0.05$) \\
\midrule
GPT-4o vs GPT-4o-mini & 0.603 & No \\
GPT-4o vs GPT-5 & $<0.001$ & Yes \\
GPT-4o vs LLaMA-3 & $<0.001$ & Yes \\
GPT-4o vs Mistral-7B & $<0.001$ & Yes \\
GPT-4o-mini vs GPT-5 & $<0.001$ & Yes \\
GPT-4o-mini vs LLaMA-3 & $<0.001$ & Yes \\
GPT-4o-mini vs Mistral-7B & $<0.001$ & Yes \\
GPT-5 vs LLaMA-3 & 0.133 & No \\
GPT-5 vs Mistral-7B & 0.077 & No \\
LLaMA-3 vs Mistral-7B & 0.002 & Yes \\
\bottomrule
\end{tabular}
\end{table}

\section{Discussion}
\label{sec:discussion}

Our findings reveal a clear mismatch between predictive performance and confidence behavior in LLMs for medical QA. While models like GPT-5 achieve high accuracy and relatively stable calibration, most models exhibit persistent overconfidence wherein reported confidence consistently exceeds actual correctness. This effect is particularly pronounced in open-weight models, e.g., LLaMA and Mistral, which frequently assign near maximal confidence to incorrect predictions. These results indicate that model generated confidence does not reliably reflect underlying epistemic uncertainty especially in complex clinical reasoning settings.

The degraded setting further highlights a fundamental limitation in uncertainty awareness. When the correct answer is removed, models rarely abstain and instead continue to select incorrect options with high confidence. This behavior reflects a strong bias toward forced answering wherein models prioritize producing an answer even in the absence of sufficient evidence. Moreover, introducing uncertainty cues at the prompt level does not meaningfully mitigate this behavior. This suggests that prompt-based approaches alone are insufficient to induce reliable uncertainty awareness or abstention behavior in current LLMs.

A plausible explanation for these observations lies in the training objectives of LLMs. Since models are optimized for next-token prediction and are predominantly trained on tasks where a valid answer exists, they implicitly learn to generate informative outputs rather than abstain. In addition, the lack of explicit supervision signals for abstention or uncertainty estimation during training may contribute to this limitation. 

These findings raise important safety concerns for the deployment of LLMs in clinical decision support systems. In high-stakes medical contexts confidently incorrect predictions may mislead clinicians and potentially lead to harmful outcomes. Therefore, improving the reliability of uncertainty estimation should be a key priority for future research. Promising directions include developing explicit abstention mechanisms, incorporating calibration aware training objectives, and designing evaluation benchmarks that systematically assess model behavior under incomplete or ambiguous clinical information. 

Overall, our results demonstrate that current LLMs, despite strong predictive performance, exhibit fundamental limitations in uncertainty awareness. Addressing these limitations is essential for ensuring safe and reliable deployment of LLMs in clinical practice.

\section{Limitations}
\label{sec:limitations}

This study has several limitations that should be considered when interpreting the results. First, our experiments are conducted on the MedMCQA dataset, which is although large and diverse, may not fully represent the complexity and variability of real world clinical decision-making. As a result, the generalizability of our findings to other datasets or clinical environments may be limited. Second, uncertainty in our framework is introduced through prompt-based linguistic cues rather than naturally occurring ambiguity. While this design enables systematic and reproducible experimentation, it may not fully reflect the uncertainty encountered in real clinical practice. Finally, our experiments are limited to a fixed set of models and a static multiple-choice setting. 
Future work should extend this framework to a broader range of models, additional medical QA benchmarks, and more realistic clinical scenarios, including open-ended reasoning tasks, interactive decision-support tasks, and multimodal clinical settings.

\section{Conclusion and Future Work}
\label{sec:conclusion}
This work presents a systematic evaluation of Large Language Models (LLMs) under conditions of clinical uncertainty and missing information. We evaluate five models across multiple dimensions, including predictive accuracy, confidence calibration, abstention behavior, hallucination rates, and unsafe confident errors. Our results reveal a fundamental limitation of current LLMs, i.e., confidence estimates do not reliably track underlying uncertainty. Across all experimental settings, models often remain highly confident even when predictive accuracy decreases or when sufficient clinical information is absent. This issue is particularly pronounced in open-weight models, which exhibit severe overconfidence, large calibration gaps, and high rates of unsafe confident errors. While proprietary models demonstrate relatively better calibration and a higher tendency to abstain, they still frequently produce confident but incorrect predictions. We further show that prompt-based linguistic uncertainty framing does not meaningfully improve model behavior. Even when uncertainty is explicitly introduced, models fail to reduce confidence or abstain appropriately. This suggests that current LLMs lack inherent mechanisms for uncertainty awareness and remain biased toward producing answers even in the absence of sufficient evidence.

These findings highlight that predictive accuracy alone is insufficient as a reliability metric for medical AI systems. Reliable deployment in clinical settings requires additional evaluation dimensions, including confidence calibration, abstention behavior, and safety oriented metrics. Future work should focus on developing calibration-aware training objectives, incorporating explicit abstention mechanisms, and designing benchmarks that systematically evaluate model behavior under incomplete or ambiguous clinical information. Ultimately, improving uncertainty awareness is essential to enable the safe and trustworthy deployment of LLMs in real-world clinical decision-making.

\section{Acknowledgment}
This work was supported by Macquarie University through the International Research Training Program (iRTP) Scholarship and the International Macquarie University Research Excellence Scholarship (iMQRES) under Scholarship Allocation Number: 20246621.

\bibliographystyle{IEEEtran}
\bibliography{references}

@article{10.1007/s10115-025-02620-1,
author = {Perera, Manoj Madushanka and Mahmood, Adnan and Wijethilake, Kasun Eranda and Islam, Fahmida and Tahermazandarani, Maryam and Sheng, Quan Z.},
title = {A Survey of the State of the Art in Conversational Question Answering Systems},
year = {2026},
issue_date = {May 2026},
publisher = {Springer-Verlag},
address = {Berlin, Heidelberg},
volume = {68},
number = {1},
issn = {0219-1377},
doi = {10.1007/s10115-025-02620-1},
journal = {Knowledge and Information Systems},
month = dec,
numpages = {50}
}

@inproceedings{guo2017calibration,
author = {Guo, Chuan and Pleiss, Geoff and Sun, Yu and Weinberger, Kilian Q.},
title = {{On Calibration of Modern Neural Networks}},
year = {2017},
publisher = {JMLR.org},
booktitle = {{Proceedings of the 34th International Conference on Machine Learning - Volume 70}},
pages = {1321–1330},
numpages = {10},
location = {Sydney, NSW, Australia},
series = {ICML'17}
}

@article{singhal2023large,
  title={{Large Language Models Encode Clinical Knowledge}},
  author={Singhal, Karan and Azizi, Shekoofeh and Tu, Tao and Mahdavi, S Sara and Wei, Jason and Chung, Hyung Won and Scales, Nathan and Tanwani, Ajay and Cole-Lewis, Heather and Pfohl, Stephen and others},
  journal={{Nature}},
  volume={620},
  number={7972},
  pages={172--180},
  year={2023},
  publisher={Nature Publishing Group UK London}
}

@article{kadavath2022language,
  title={{Language Models (mostly) Know What They Know}},
  author={Kadavath, Saurav and Conerly, Tom and Askell, Amanda and Henighan, Tom and Drain, Dawn and Perez, Ethan and Schiefer, Nicholas and Hatfield-Dodds, Zac and DasSarma, Nova and Tran-Johnson, Eli and others},
  journal={arXiv preprint arXiv:2207.05221},
  year={2022}
}

@article{ji2023survey,
  title={{Survey of Hallucination in Natural Language Generation}},
  author={Ji, Ziwei and Lee, Nayeon and Frieske, Rita and Yu, Tiezheng and Su, Dan and Xu, Yan and Ishii, Etsuko and Bang, Ye Jin and Madotto, Andrea and Fung, Pascale},
  journal={{ACM Computing Surveys}},
  volume={55},
  number={12},
  pages={1--38},
  year={2023},
  publisher={ACM New York, NY}
}

@article{huang2025survey,
  title={{A Survey on Hallucination in Large Language Models: Principles, Taxonomy, Challenges, and Open Questions}},
  author={Huang, Lei and Yu, Weijiang and Ma, Weitao and Zhong, Weihong and Feng, Zhangyin and Wang, Haotian and Chen, Qianglong and Peng, Weihua and Feng, Xiaocheng and Qin, Bing and others},
  journal={{ACM Transactions on Information Systems}},
  volume={43},
  number={2},
  pages={1--55},
  year={2025},
  publisher={ACM New York, NY}
}

@article{brown2020language,
  title={{Language Models Are Few-shot Learners}},
  author={Brown, Tom and Mann, Benjamin and Ryder, Nick and Subbiah, Melanie and Kaplan, Jared D and Dhariwal, Prafulla and Neelakantan, Arvind and Shyam, Pranav and Sastry, Girish and Askell, Amanda and others},
  journal={{Advances in Neural Information Processing Systems}},
  volume={33},
  pages={1877--1901},
  year={2020}
}

@inproceedings{yin2023large,
  title={{Do Large Language Models Know What They Don’t Know?}},
  author={Yin, Zhangyue and Sun, Qiushi and Guo, Qipeng and Wu, Jiawen and Qiu, Xipeng and Huang, Xuan-Jing},
  booktitle={{Findings of the Association for Computational Linguistics: ACL 2023}},
  pages={8653--8665},
  year={2023}
}

@article{steyvers2025large,
  title={{What Large Language Models Know and What People Think They Know}},
  author={Steyvers, Mark and Tejeda, Heliodoro and Kumar, Aakriti and Belem, Catarina and Karny, Sheer and Hu, Xinyue and Mayer, Lukas W and Smyth, Padhraic},
  journal={{Nature Machine Intelligence}},
  volume={7},
  number={2},
  pages={221--231},
  year={2025},
  publisher={{Nature Publishing Group UK London}}
}

@article{kung2023performance,
  title={{Performance of ChatGPT on USMLE: Potential for AI-Assisted Medical Education using Large Language Models}},
  author={Kung, Tiffany H and Cheatham, Morgan and Medenilla, Arielle and Sillos, Czarina and De Leon, Lorie and Elepa{\~n}o, Camille and Madriaga, Maria and Aggabao, Rimel and Diaz-Candido, Giezel and Maningo, James and others},
  journal={{PLoS Digital Health}},
  volume={2},
  number={2},
  pages={e0000198},
  year={2023},
  publisher={{Public Library of Science}}
}

@inproceedings{pal2022medmcqa,
  title={{MedMCQA: A large-Scale Multi-Subject Multi-Choice Dataset for Medical Domain Question Answering}},
  author={Pal, Ankit and Umapathi, Logesh Kumar and Sankarasubbu, Malaikannan},
  booktitle={{Conference on Health, Inference, and Learning}},
  pages={248--260},
  year={2022},
  organization={PMLR}
}

@inproceedings{tripathi2025confidence,
  title={{The Confidence Paradox: Can LLM Know When It’s Wrong?}},
  author={Tripathi, Sahil and Nafis, Md Tabrez and Hussain, Imran and Gao, Jiechao},
  booktitle={Proceedings of the 14th International Joint Conference on Natural Language Processing and the 4th Conference of the Asia-Pacific Chapter of the Association for Computational Linguistics},
  pages={2078--2087},
  year={2025}
}

@inproceedings{xu2025language,
  title={{Do Language Models Mirror Human Confidence? Exploring Psychological Insights to Address Overconfidence in LLMs}},
  author={Xu, Chenjun and Wen, Bingbing and Han, Bin and Wolfe, Robert and Wang, Lucy Lu and Howe, Bill},
  booktitle={Findings of the Association for Computational Linguistics: ACL 2025},
  pages={25655--25672},
  year={2025}
}

@inproceedings{groot2024overconfidence,
  title={{Overconfidence is Key: Verbalized Uncertainty Evaluation in Large Language and Vision-Language Models}},
  author={Groot, Tobias and Valdenegro-Toro, Matias},
  booktitle={{Proceedings of the 4th Workshop on Trustworthy Natural Language Processing (TrustNLP 2024)}},
  pages={145--171},
  year={2024}
}

@article{bentegeac2025token,
  title={{Token Probabilities to Mitigate Large Language Models Overconfidence in Answering Medical Questions: Quantitative Study}},
  author={Bentegeac, Rapha{\"e}l and Le Guellec, Bastien and Kuchcinski, Gr{\'e}gory and Amouyel, Philippe and Hamroun, Aghiles},
  journal={{Journal of Medical Internet Research}},
  volume={27},
  pages={e64348},
  year={2025},
  publisher={JMIR Publications Toronto, Canada}
}

@inproceedings{tian2023just,
  title={{Just Ask for Calibration: Strategies for Eliciting Calibrated Confidence Scores from Language Models Fine-Tuned with Human Feedback}},
  author={Tian, Katherine and Mitchell, Eric and Zhou, Allan and Sharma, Archit and Rafailov, Rafael and Yao, Huaxiu and Finn, Chelsea and Manning, Christopher D},
  booktitle={{Proceedings of the 2023 Conference on Empirical Methods in Natural Language Processing}},
  pages={5433--5442},
  year={2023}
}

@article{qiu2024semantic,
  title={{Semantic Density: Uncertainty Quantification for Large Language Models Through Confidence Measurement in Semantic Space}},
  author={Qiu, Xin and Miikkulainen, Risto},
  journal={{Advances in Neural Information Processing Systems}},
  volume={37},
  pages={134507--134533},
  year={2024}
}

@inproceedings{wightman2023strength,
  title={{Strength in Numbers: Estimating Confidence of Large Language Models by Prompt Agreement}},
  author={Wightman, Gwenyth Portillo and Delucia, Alexandra and Dredze, Mark},
  booktitle={{Proceedings of the 3rd Workshop on Trustworthy Natural Language Processing (TrustNLP 2023)}},
  pages={326--362},
  year={2023}
}

@inproceedings{madhusudhan2025llms,
  title={{Do LLMs Know When to Not Answer? Investigating Abstention Abilities of Large Language Models}},
  author={Madhusudhan, Nishanth and Madhusudhan, Sathwik Tejaswi and Yadav, Vikas and Hashemi, Masoud},
  booktitle={{Proceedings of the 31st International Conference on Computational Linguistics}},
  pages={9329--9345},
  year={2025}
}

@article{mcnemar1947note,
  title={{Note on the Sampling Error of the Difference between Correlated Proportions or Percentages}},
  author={McNemar, Quinn},
  journal={Psychometrika},
  volume={12},
  number={2},
  pages={153--157},
  year={1947},
  publisher={Springer-Verlag}
}

@article{wilcoxon1945individual,
  title={{Individual Comparisons by Ranking Methods}},
  author={Wilcoxon, Frank},
  journal={{Biometrics Bulletin}},
  volume={1},
  number={6},
  pages={80--83},
  year={1945},
  publisher={JSTOR}
}

@article{holm1979simple,
  title={{A Simple Sequentially Rejective Multiple Test Procedure}},
  author={Holm, Sture},
  journal={{Scandinavian Journal of Statistics}},
  pages={65--70},
  year={1979},
  publisher={JSTOR}
}

@inproceedings{geng2024survey,
  title={{A Survey of Confidence Estimation and Calibration in Large Language Models}},
  author={Geng, Jiahui and Cai, Fengyu and Wang, Yuxia and Koeppl, Heinz and Nakov, Preslav and Gurevych, Iryna},
  booktitle={Proceedings of the 2024 Conference of the North American Chapter of the Association for Computational Linguistics: Human Language Technologies (Volume 1: Long Papers)},
  pages={6577--6595},
  year={2024}
}

@article{chhikara2025mind,
  title={{Mind the Confidence Gap: Overconfidence, Calibration, and Distractor Effects in Large Language Models}},
  author={Chhikara, Prateek},
  journal={arXiv preprint arXiv:2502.11028},
  year={2025}
}

@article{machcha2026knowing,
  title={{Knowing When to Abstain: Medical LLMs Under Clinical Uncertainty}},
  author={Machcha, Sravanthi and Yerra, Sushrita and Gupta, Sahil and Sahoo, Aishwarya and Sultana, Sharmin and Yu, Hong and Yao, Zonghai},
  journal={arXiv preprint arXiv:2601.12471},
  year={2026}
}

\end{document}